\documentclass[number,sort&compress]{elsarticle}

\usepackage{amsmath}
\usepackage{amsfonts}
\usepackage{amssymb}
\usepackage{amsthm}
\usepackage{graphicx}
\usepackage{multirow}
\usepackage[english]{babel}
\usepackage{natbib}
\usepackage{booktabs}

\newcommand{\Prob}{\operatorname{P}}
\def\bbbone{{\mathchoice {\rm 1\mskip-4mu l} {\rm 1\mskip-4mu l}
  {\rm 1\mskip-4.5mu l} {\rm 1\mskip-5mu l}}}

\newdefinition{example}{Example}

\begin{document}

\title{Identifying Significant Edges in Graphical Models of Molecular Networks}
\author[marco]{Marco Scutari\corref{cor1}}
\ead{m.scutari@ucl.ac.uk}

\author[radha]{Radhakrishnan Nagarajan}
\ead{rnagarajan@uky.edu}

\address[marco]{Genetics Institute, University College London, \linebreak 
  Darwin Building, Gower Street, WC1E 6BT, London, United Kingdom.}
\address[radha]{Division of Biomedical Informatics, Department of Biostatistics, \linebreak
  College of Public Health, University of Kentucky, 725 Rose Street, \linebreak 
  Multidisciplinary Science Bldg, 230F, Lexington, KY 40536-0082, USA.}

\begin{abstract}

\textit{Objective:} Modelling the associations from high-throughput experimental
molecular data has provided unprecedented insights into biological pathways and
signalling mechanisms. Graphical models and networks have especially proven to
be useful abstractions in this regard. Ad-hoc thresholds are often used in 
conjunction with structure learning algorithms to determine significant 
associations. The present study overcomes this limitation by proposing a 
statistically-motivated approach for identifying significant associations in
a network.

\textit{Methods and Materials:} A new method that identifies significant 
associations in graphical models by estimating the threshold minimising
the $L_{\mathrm{1}}$ norm between the cumulative distribution function (CDF)
of the observed edge confidences and those of its asymptotic counterpart is
proposed. The effectiveness of the proposed method is demonstrated on popular
synthetic data sets as well as publicly available experimental molecular data
corresponding to gene and protein expression profiles.

\textit{Results:} The improved performance of the proposed approach is
demonstrated across the synthetic data sets using sensitivity, specificity
and accuracy as performance metrics. The results are also demonstrated 
across varying sample sizes and three different structure learning algorithms
with widely varying assumptions. In all cases, the proposed approach has
specificity and accuracy close to $1$, while sensitivity increases linearly
in the logarithm of the sample size. The estimated threshold systematically
outperforms common ad-hoc ones in terms of sensitivity while maintaining
comparable levels of specificity and accuracy. Networks from experimental
data sets are  reconstructed accurately with respect to the results from
the original papers.

\textit{Conclusion:} Current studies use structure learning algorithms in
conjunction with ad-hoc thresholds for identifying significant associations
in graphical abstractions of biological pathways and signalling mechanisms.
Such an ad-hoc choice can have pronounced effect on attributing biological
significance to the associations in the resulting network and possible
downstream analysis. The statistically-motivated approach presented in
this study has been shown to outperform ad-hoc thresholds and is expected to
alleviate spurious conclusions of significant associations in such graphical
abstractions.

\end{abstract}


\maketitle

\section{Introduction and background}

Graphical models \cite{koller,pearl} are a class of statistical models which
combine the rigour of a probabilistic approach with the intuitive representation
of relationships given by graphs. They are composed by a set $\mathbf{X} = \{X_1,
X_2, \ldots, X_N\}$ of \emph{random variables} describing the quantities of
interest and a \emph{graph} $\mathcal{G} = (\mathbf{V}, E)$ in which each
\emph{node} or \emph{vertex} $v \in \mathbf{V}$ is associated with one of the
random variables in $\mathbf{X}$ (they are usually referred to interchangeably).
The \emph{edges} $e \in E$ are used to express the dependence relationships
among the variables in $\mathbf{X}$. The set of these relationships is often
referred to as the \emph{dependence structure} of the graph. Different classes
of graphs express these relationships with different semantics, which have in
common the principle that graphical separation of two vertices implies the
conditional independence of the corresponding random variables \cite{pearl}.
The two examples most commonly found in literature are \emph{Markov networks}
\cite{whittaker,edwards}, which use undirected graphs, and \emph{Bayesian
networks} (BNs) \cite{neapolitan,korb}, which use directed acyclic graphs.

In principle, there are many possible choices for the joint distribution of 
$\mathbf{X}$, depending on the nature of the data. However, literature have
focused mostly on two cases: the \textit{discrete case} \cite{whittaker,heckerman},
in which both $\mathbf{X}$ and the $X_i$ are multinomial random variables,
and the \textit{continuous case} \cite{whittaker,heckerman3}, in which
$\mathbf{X}$ is multivariate normal and the $X_i$ are univariate normal random
variables. In the former, the parameters of interest are the \textit{conditional
probabilities} associated with each variable, usually represented as conditional
probability tables; in the latter, the parameters of interest are the
\textit{partial correlation coefficients} between each variable and its
neighbours (i.e. the adjacent nodes in $\mathcal{G}$).

The estimation of the structure of the graph $\mathcal{G}$ is called
\emph{structure learning} \cite{koller,edwards}, and involves determining the
graph structure that encodes the conditional independencies present in the data.
Ideally it should coincide with the dependence structure of $\mathbf{X}$, or it
should at least identify a distribution as close as possible to the correct one
in the probability space. Several algorithms have been presented in literature
for this problem, thanks to the application of many results from probability,
information and optimisation theory. Despite differences in theoretical
backgrounds and terminology, they can all be grouped into only three classes:
\textit{constraint-based algorithms}, that are based on conditional independence
tests; \textit{score-based algorithms}, that are based on goodness-of-fit scores;
and \textit{hybrid algorithms}, that combine the previous two approaches. For
some examples see Bromberg et al. \cite{gsmn}, Castelo and Roverato \cite{roverato},
Friedman et al. \cite{sc}, Larra{\~n}aga et al. \cite{larranaga} and Tsamardinos
et al. \cite{mmhc}.

On the other hand, the development of techniques for assessing the statistical
robustness of network structures learned from data (e.g. the presence of artefacts
arising from noisy data) has been limited. Structure learning algorithms are
commonly studied measuring differences from the true (known) structure of a small
number of reference data sets \cite{bnrepository,uci}. The usefulness of such an
approach in investigating networks learned from real-world data sets is limited,
since the true structure of their probability distribution is unknown.

A more systematic approach to model assessment, and in particular to the problem
of identifying statistically significant features in a network, has been developed
by Friedman et al. \cite{friedman} using bootstrap resampling \cite{efron} and
model averaging \cite{claeskens}.
It can be summarised as follows:
\begin{enumerate}
  \item For $b = 1, 2, \ldots, m$:
    \begin{enumerate}
      \item sample a new data set $\mathbf{X}^*_b$ from the original data
        $\mathbf{X}$ using either parametric or nonparametric bootstrap;
      \item learn the structure of the graphical model $\mathcal{G}_b =
        (\mathbf{V}, E_b)$ from $\mathbf{X}^*_b$.
    \end{enumerate}
  \item Estimate the probability that each possible edge $e_i$, $i = 1, \ldots, k$
    is present in the true network structure $\mathcal{G}_0 = (\mathbf{V}, E_0)$ as
    \begin{equation}
      \label{eq:bootconf}
      \hat\Prob(e_i) = \frac{1}{m} \sum_{b=1}^m \bbbone_{\{e_i \in E_b\}},
    \end{equation}
    where $\bbbone_{{\{e_i \in E_b\}}}$ is the indicator function of the event
    $\{e_i \in E_b\}$ (i.e., it is equal to $1$ if $e_i \in E_b$ and $0$
    otherwise).
\end{enumerate}
The empirical probabilities $\hat\Prob(e_i)$ are known as \emph{edge intensities}
or \emph{arc strengths}, and can be interpreted as the degree of \emph{confidence}
that $e_i$ is present in the network structure $\mathcal{G}_0$ describing the
true dependence structure of $\mathbf{X}$\footnote{The probabilities $\hat\Prob(e_i)$
are in fact an estimator of the expected value of the $\{0,1\}$ random vector
describing the presence of each possible edge in $\mathcal{G}_0$. As such, they
do not sum to one and are dependent on one another in a nontrivial way.}. However,
they are difficult to evaluate, because the probability distribution of the networks
$\mathcal{G}_b$ in the space of the network structures is unknown. As a result,
the value of the confidence threshold (i.e. the minimum degree of confidence for
an edge to be significant and therefore accepted as an edge of $\mathcal{G}_0$)
is an unknown function of both the data and the structure learning algorithm. This
is a serious limitation in the identification of significant edges and has led
to the use of ad-hoc, pre-defined thresholds in spite of the impact on model
assessment evidenced by several studies \cite{friedman,husmeier}. An exception
is Nagarajan et al. \cite{myogenic}, whose approach will be discussed below.

Apart from this limitation, Friedman's approach is very general and can be used
in a wide range of settings. First of all, it can be applied to any kind of
graphical model with only minor adjustments (for example, accounting for the
direction of the edges in BNs, see Sec. \ref{sec:geneprof}).
No distributional assumption on the data is required in addition to the ones
needed by the structure learning algorithm. No assumption is made on the
latter, either, so any score-based, constraint-based or hybrid algorithm can
be used. Furthermore, parallel computing can easily be used to offset the
additional computational complexity introduced by model averaging, because
bootstrap is embarrassingly parallel.

In this paper, we propose a statistically-motivated estimator for the confidence
threshold minimising the $L_{\mathrm{1}}$ norm between the cumulative distribution
function (CDF) of the observed confidence levels and the CDF of the confidence 
levels of the unknown network $\mathcal{G}_0$. Subsequently, we demonstrate the
effectiveness of the proposed approach by re-investigating two experimental data
sets from Nagarajan et al. \cite{myogenic} and Sachs et al. \cite{sachs}.

\section{Selecting significant edges}
\label{sec:approach}

Consider the empirical probabilities $\hat\Prob(e_i)$ defined in Eq. \ref{eq:bootconf},
and denote them with $\mathbf{\hat{p}} = \{\hat{p}_i, i = 1, \ldots, k\}$. For a
graph with $N$ nodes, $k = N(N - 1)/2$. Furthermore, consider the order statistic 
\begin{align}
  &\mathbf{\hat{p}_{(\cdot)}} = \left(\hat{p}_{(1)}, \hat{p}_{(2)}, \ldots, \hat{p}_{(k)}\right)&
  &\text{with}&
  &\hat{p}_{(1)} \leqslant \hat{p}_{(2)} \leqslant \ldots \leqslant \hat{p}_{(k)}
\end{align}
derived from $\mathbf{\hat{p}}$. It is intuitively clear that the first elements of
$\mathbf{\hat{p}_{(\cdot)}}$ are more likely to be associated with non-significant
edges, and that the last elements of $\mathbf{\hat{p}_{(\cdot)}}$ are more likely
to be associated with significant edges. The ideal configuration $\mathbf{\tilde{p}_{(\cdot)}}$
of $\mathbf{\hat{p}_{(\cdot)}}$ would be 
\begin{equation}
  \tilde{p}_{(i)} = \left\{
    \begin{aligned}
      &1& &\text{if $e_{(i)} \in E_0$}     \\
      &0& &\text{otherwise}&
    \end{aligned}
    \right.,
\end{equation}
that is the set of probabilities that characterises any edge as either significant
or non-significant without any uncertainty. In other words, 
\begin{equation}
  \mathbf{\tilde{p}_{(\cdot)}} = \{0, \ldots, 0, 1, \ldots, 1\}.
\end{equation}
Such a configuration arises from the limit case in which all the networks
$\mathcal{G}_b$ have exactly the same structure. This may happen in practice with
a consistent structure learning algorithm when the sample size is large
\cite{lauritzen,ges}.

A useful characterisation of $\mathbf{\hat{p}_{(\cdot)}}$ and $\mathbf{\tilde{p}_{(\cdot)}}$
can be obtained through the empirical CDFs of the respective elements,
\begin{equation}
  F_{\mathbf{\hat{p}_{(\cdot)}}}(x) = \frac{1}{k}\sum_{i = 1}^k \bbbone_{\{\hat{p}_{(i)} < x\}} 
\end{equation}
and 
\begin{equation}
  F_{\mathbf{\tilde{p}_{(\cdot)}}}(x) = \left\{
    \begin{aligned}
      &0& &\text{if $x \in (-\infty, 0)$} \\
      &t& &\text{if $x \in \left[0, 1\right)$}     \\
      &1& &\text{if $x \in [1, +\infty) $}&
    \end{aligned}
    \right..
\end{equation}
In particular, $t$ corresponds to the fraction of elements of $\mathbf{\tilde{p}_{(\cdot)}}$
equal to zero and is a measure of the fraction of non-significant edges. 
At the same time, $t$ provides a threshold for separating the elements of
$\mathbf{\tilde{p}_{(\cdot)}}$, namely
\begin{equation}
  e_{(i)} \in E_0 \Longleftrightarrow \tilde{p}_{(i)} > F^{-1}_{\mathbf{\tilde{p}_{(\cdot)}}}(t)
\end{equation}
where $F^{-1}_{\mathbf{\tilde{p}_{(\cdot)}}}(t) = \inf_{x \in \mathbb{R}} 
\left\{ F_{\mathbf{\tilde{p}_{(\cdot)}}}(x) \geqslant t\right\}$ is the
\textit{quantile function} \cite{degroot}.

More importantly, estimating $t$ from data provides a statistically motivated
threshold for separating significant edges from non-significant ones. In
practice, this amounts to approximating the ideal, asymptotic empirical CDF
$F_{\mathbf{\tilde{p}_{(\cdot)}}}$ with its finite sample estimate
$F_{\mathbf{\hat{p}_{(\cdot)}}}$. Such an approximation can be computed in many
different ways, depending on the norm used to measure the distance between
$F_{\mathbf{\hat{p}_{(\cdot)}}}$ and $F_{\mathbf{\tilde{p}_{(\cdot)}}}$ as a
function of $t$. Common choices are the $L_\mathrm{p}$ family of norms \cite{lp},
which includes the Euclidean norm, and Csiszar's $f$-divergences \cite{csiszar},
which include Kullback-Leibler divergence.

\begin{figure}[t]
  \includegraphics[width=\textwidth]{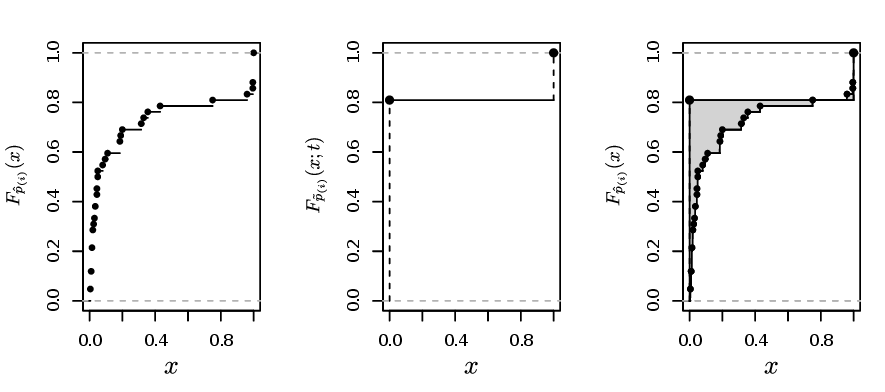}
  \caption{The empirical CDF $F_{\mathbf{\hat{p}_{(\cdot)}}}$ (left), the CDF
    $F_{\mathbf{\tilde{p}_{(\cdot)}}}$ (centre) and the $L_1$ norm between the
    two (right), shaded in grey.}
  \label{fig:ecdf}
\end{figure}

The $L_1$ norm
\begin{equation}
\label{eq:l1}
  L_{1}\left(t; \mathbf{\hat{p}_{(\cdot)}}\right) =
    \int \left| F_{\mathbf{\hat{p}_{(\cdot)}}}(x) - F_{\mathbf{\tilde{p}_{(\cdot)}}}(x; t) \right| dx
\end{equation}
appears to be particularly suited to this problem; an example is shown in Fig.
\ref{fig:ecdf}. First of all, note that $F_{\mathbf{\hat{p}_{(\cdot)}}}$ is
piecewise constant, changing value only at the points $\hat{p}_{(i)}$; this
descends from the definition of empirical CDF. Therefore, for the problem at 
hand Eq. \ref{eq:l1} simplifies to 
\begin{equation}
  L_{\mathrm{1}}\left(t; \mathbf{\hat{p}_{(\cdot)}}\right) =
    \sum_{x_i \in \left\{\{0\} \cup \mathbf{\hat{p}_{(\cdot)}} \cup \{1\}\right\}}
      \left| F_{\mathbf{\hat{p}_{(\cdot)}}}(x_i) - t \right| (x_{i+1} - x_i),
\end{equation}
which can be computed in linear time from $\mathbf{\hat{p}_{(\cdot)}}$. Its
minimisation is also straightforward using linear programming \cite{nocedal}. 
Furthermore, compared to the more common $L_2$ norm
\begin{equation}
  L_{2}\left(t; \mathbf{\hat{p}_{(\cdot)}}\right) =
    \int \left[ F_{\mathbf{\hat{p}_{(\cdot)}}}(x) - F_{\mathbf{\tilde{p}_{(\cdot)}}}(x; t) \right]^2 dx
\end{equation}
or the $L_\infty$ norm 
\begin{equation}
  L_{\infty}\left(t; \mathbf{\hat{p}_{(\cdot)}}\right) =
    \max_{x \in [0, 1]} \left\{ \left|F_{\mathbf{\hat{p}_{(\cdot)}}}(x) - F_{\mathbf{\tilde{p}_{(\cdot)}}}(x; t)\right| \right\},
\end{equation}
the $L_1$ norm does not place as much weight on large deviations compared to
small ones, making it robust against a wide variety of configurations of
$\mathbf{\hat{p}_{(\cdot)}}$. 

Then the identification of significant edges can be thought of either as a
\emph{least absolute deviations estimation} or an \emph{$L_1$ approximation}
of the form
\begin{equation}
  \hat{t} = \underset{t \in [0, 1]}{\operatorname{argmin}} \; L_{1}\left(t; \mathbf{\hat{p}_{(\cdot)}}\right)
\end{equation}
followed by the application of the following rule:
\begin{equation}
  e_{(i)} \in E_0 \Longleftrightarrow \hat{p}_{(i)} > F^{-1}_{\mathbf{\hat{p}_{(\cdot)}}}(\hat{t}).
\end{equation}
Note that, even though edges are individually identified as as significant or
non-significant, they are not identified independently of each other because
$\hat t$ is a function of the whole $\mathbf{\hat{p}_{(\cdot)}}$. 

A simple example is illustrated below.

\begin{example}
\label{ex}
  Consider a graphical model based on an undirected graph $\mathcal{G}$ with
  node set $\mathbf{V} = \{A, B, C, D\}$. The set of possible edges of 
  $\mathcal{G}$ contains $6$ elements: $(A, B)$, $(A, C)$, $(A, D)$, $(B, C)$,
  $(B, D)$ and $(C, D)$. Suppose that that we have estimated the following
  confidence values: 
  \begin{align}
    &\hat{p}_{AB} = 0.2242,& &\hat{p}_{AC} = 0.0460,& &\hat{p}_{AD} = 0.8935, \notag \\
    &\hat{p}_{BC} = 0.3921,& &\hat{p}_{BD} = 0.7689,& &\hat{p}_{CD} = 0.9439.
  \end{align}
  Then $\mathbf{\hat{p}_{(\cdot)}} = \{0.0460, 0.2242, 0.3921, 0.7689, 0.8935, 0.9439\}$
  and
  \begin{equation}
    F_{\mathbf{\hat{p}_{(\cdot)}}}(x) = \left\{
      \begin{aligned}
        &0& &\text{if $x \in (-\infty, 0.0460)$} \\
        &\frac{1}{6}& &\text{if $x \in \left[0.0460, 0.2242\right)$}     \\
        &\frac{2}{6}& &\text{if $x \in \left[0.2242, 0.3921\right)$}     \\
        &\frac{3}{6}& &\text{if $x \in \left[0.3921, 0.7689\right)$}     \\
        &\frac{4}{6}& &\text{if $x \in \left[0.7689, 0.8935\right)$}     \\
        &\frac{5}{6}& &\text{if $x \in \left[0.8935, 0.9439\right)$}     \\
        &1& &\text{if $x \in [0.9439, +\infty) $}&
      \end{aligned}
      \right..
  \end{equation}

  \begin{figure}[t]
    \includegraphics[width=\textwidth]{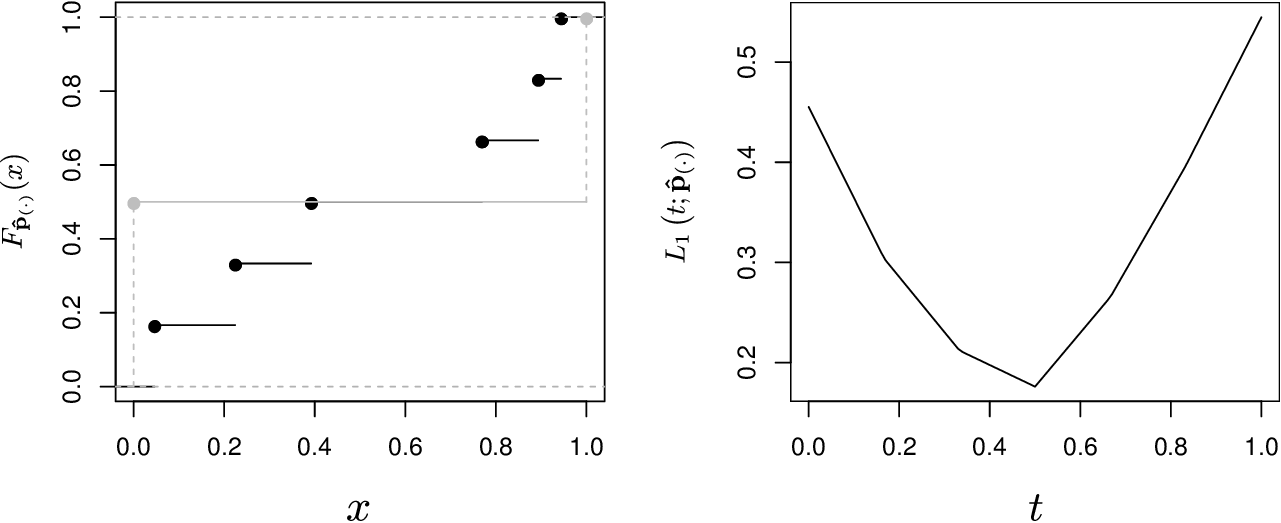}
    \caption{The CDFs $F_{\mathbf{\hat{p}_{(\cdot)}}}$ and 
      $F_{\mathbf{\tilde{p}_{(\cdot)}}}(\hat{t})$, respectively in black and
      grey (left), and the $L_{1}\left(t; \mathbf{\hat{p}_{(\cdot)}}\right)$
      norm (right) from Example \ref{ex}.}
    \label{fig:example}
  \end{figure}

  The $L_1$ norm takes the form
  \begin{multline}
    L_{1}\left(t; \mathbf{\hat{p}_{(\cdot)}}\right) = |0 - t|(0.0460 - 0) +
      \left|\frac{1}{6} - t\right|(0.2242 - 0.0460) + \\
      \left|\frac{2}{6} - t\right|(0.3921 - 0.2242) + 
      \left|\frac{3}{6} - t\right|(0.7689 - 0.3921) + \\
      \left|\frac{4}{6} - t\right|(0.8935 - 0.7689) +
      \left|\frac{5}{6} - t\right|(0.9439 - 0.8935) + \\
      \left|1 - t\right|(1 - 0.9439)
  \end{multline}
  and is minimised for $\hat{t} = 0.4999816$. Therefore, an edge is deemed
  significant if its confidence is strictly greater than 
  $F^{-1}_{\mathbf{\hat{p}_{(\cdot)}}}(0.4999816) = 0.3921$, or, equivalently,
  if it has confidence of at least $0.7689$; only $(A, D)$, $(B, D)$ and
  $(C, D)$ satisfy this condition.
\end{example}

\section{Simulation results}

We tested the proposed approach on synthetic data sets using three established
performance measures: \emph{sensitivity}, \emph{specificity} and \emph{accuracy}.
\emph{Sensitivity} is given by the proportion of edges of the true network
structure that have been correctly identified as significant. \emph{Specificity}
is given by the proportion of the edges missing from the true network structure
that have been correctly identified as non-significant. \emph{Accuracy} is given
by the proportion of edges correctly identified as either significant or
non-significant over the set of all possible edges. To that end, we generated
$400$ data sets of varying sizes ($100$, $200$, $500$, $1000$, $2000$, $5000$,
$10000$ and $20000$) from three discrete BNs commonly used as benchmarks:
\begin{itemize}
  \item the ALARM network \cite{alarm}, a network designed to provide an alarm
    message system for intensive care unit patient monitoring. Its true structure 
    is composed by $37$ nodes and $46$ edges (of $666$ possible edges), and its
    probability distribution has $509$ parameters;
  \item the HAILFINDER network \cite{hailfinder}, a network designed to forecast
    severe summer hail in northeastern Colorado. Its true structure is
    composed by $56$ nodes and $66$ edges (of $1540$ possible edges), and its
    probability distribution has $2656$ parameters;
  \item the INSURANCE network \cite{insurance}, a network designed to evaluate
    car insurance risks. Its true structure is composed by $27$ nodes and $52$
    edges (of $351$ possible edges), and its probability distribution has $984$
    parameters.
\end{itemize}
Three different structure learning algorithms were considered:
\begin{itemize}
  \item the Incremental Association Markov Blanket (IAMB) constraint-based 
    algorithm \cite{iamb}. IAMB was used to learn the Markov blanket of each
    node as a preliminary step to reduce the number of its candidate parents
    and children; a network structure satisfying these constraints is then
    identified as in the Grow-Shrink algorithm \cite{gs}. Conditional
    independence tests were performed using a shrinkage mutual information
    test \citep{mishrink} with $\alpha = 0.05$. Such a test, unlike the more
    common asymptotic $\chi^2$ mutual information test, is valid and has been
    shown to work reliably even on small samples.  An $\alpha = 0.01$ was also
    considered; however, the results were not significantly different from
    $\alpha = 0.05$ and will not be discussed separately in this paper;
  \item the Hill Climbing (HC) score-based algorithm with the Bayesian Dirichlet
    equivalent uniform (BDeu) score function, the posterior distribution of the
    network structure arising from a uniform prior distribution \cite{heckerman}.
    The equivalent sample size was set to $10$. This is the same approach
    detailed in Friedman et al. \cite{friedman}, although they considered only
    $100$ (instead of $500$) bootstrap samples for each scenario;
  \item the Max-Min Hill Climbing (MMHC) hybrid algorithm \cite{mmhc}, which
    combines the Max-Min Parents and Children (MMPC) and HC. The conditional
    independence test used in MMPC and the score functions used in HC are the
    ones illustrated in the previous points.
\end{itemize}
The performance measures were estimated for each combination of network, sample
size and structure learning algorithm as follows:
\begin{enumerate}
  \item a sample of the appropriate size was generated from either the ALARM,
    the HAILFINDER or the INSURANCE network;
  \item we estimated the confidence values $\mathbf{\hat{p}}$ for all possible
    edges from $200$ and $500$ nonparametric bootstrap samples. Since results
    are very similar, they will be discussed together;
  \item we estimated the confidence threshold $\hat t$, and identified
    significant and non-significant edges in the network. Note that the direction
    of the edges present in the network structure is effectively ignored,
    because the proposed approach focuses only those edges' presence. 
    Significant edges were then used to build an averaged network structure;
  \item we computed sensitivity, specificity and accuracy comparing the
    averaged network structure to the true one, which is known from literature.
\end{enumerate}
These steps were repeated $50$ times in order to estimate both the performance
measures and their variability. 

\begin{figure}[p]
\begin{center}
  \includegraphics[width=0.7\textwidth]{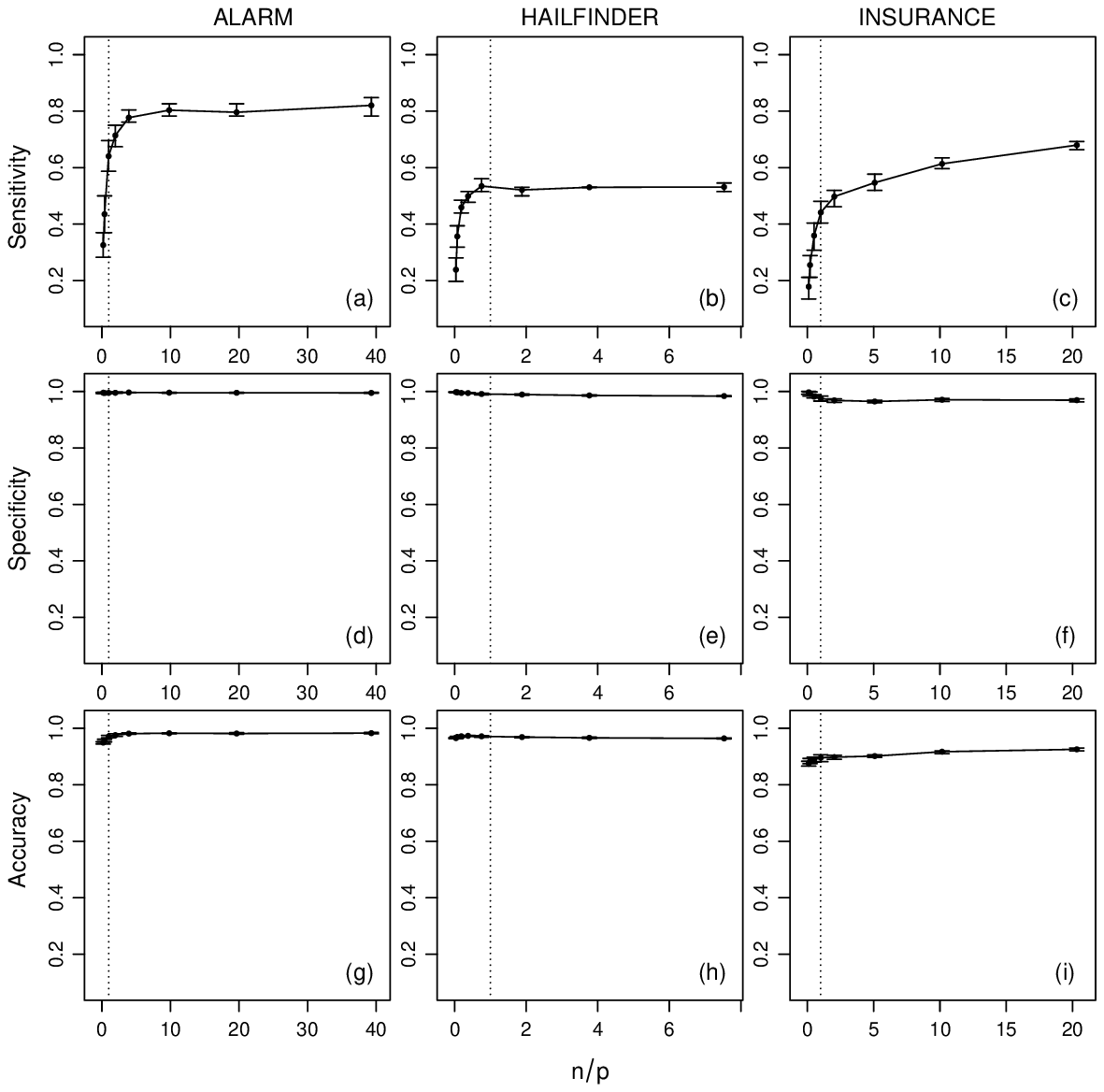}
  \caption{Average sensitivity, specificity and accuracy of IAMB for the ALARM,
    HAILFINDER and INSURANCE networks over $n/p$. Bars represent 95\% confidence
    intervals, and the dotted vertical line is $n = p$.}
  \label{fig:roc-iamb}
  \vfill
  \includegraphics[width=0.7\textwidth]{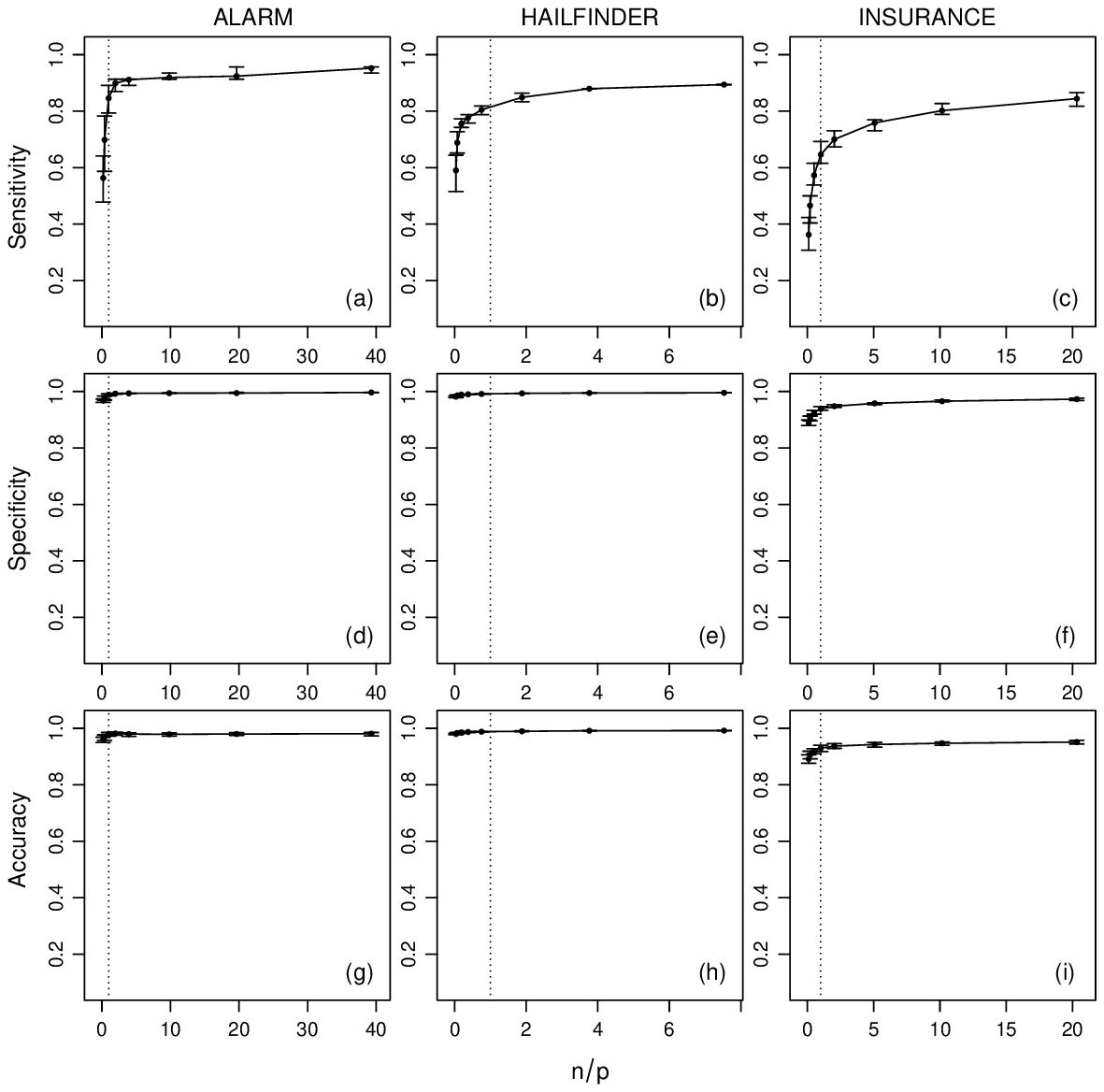}
  \caption{Average sensitivity, specificity and accuracy of HC for the ALARM, 
    HAILFINDER and INSURANCE networks over $n/p$. Bars represent 95\% confidence 
    intervals, and the dotted vertical line is $n = p$.}
  \label{fig:roc-hc}
\end{center}
\end{figure}

All the simulations and the thresholds estimation were performed with the
bnlearn package \cite{bnlearn,jss09} for R \cite{R}, which implements several
methods for structure learning, parameter estimation and inference on BNs
(including the approach proposed in Sec. \ref{sec:approach}).

\begin{figure}[t!]
\begin{center}
  \includegraphics[width=0.7\textwidth]{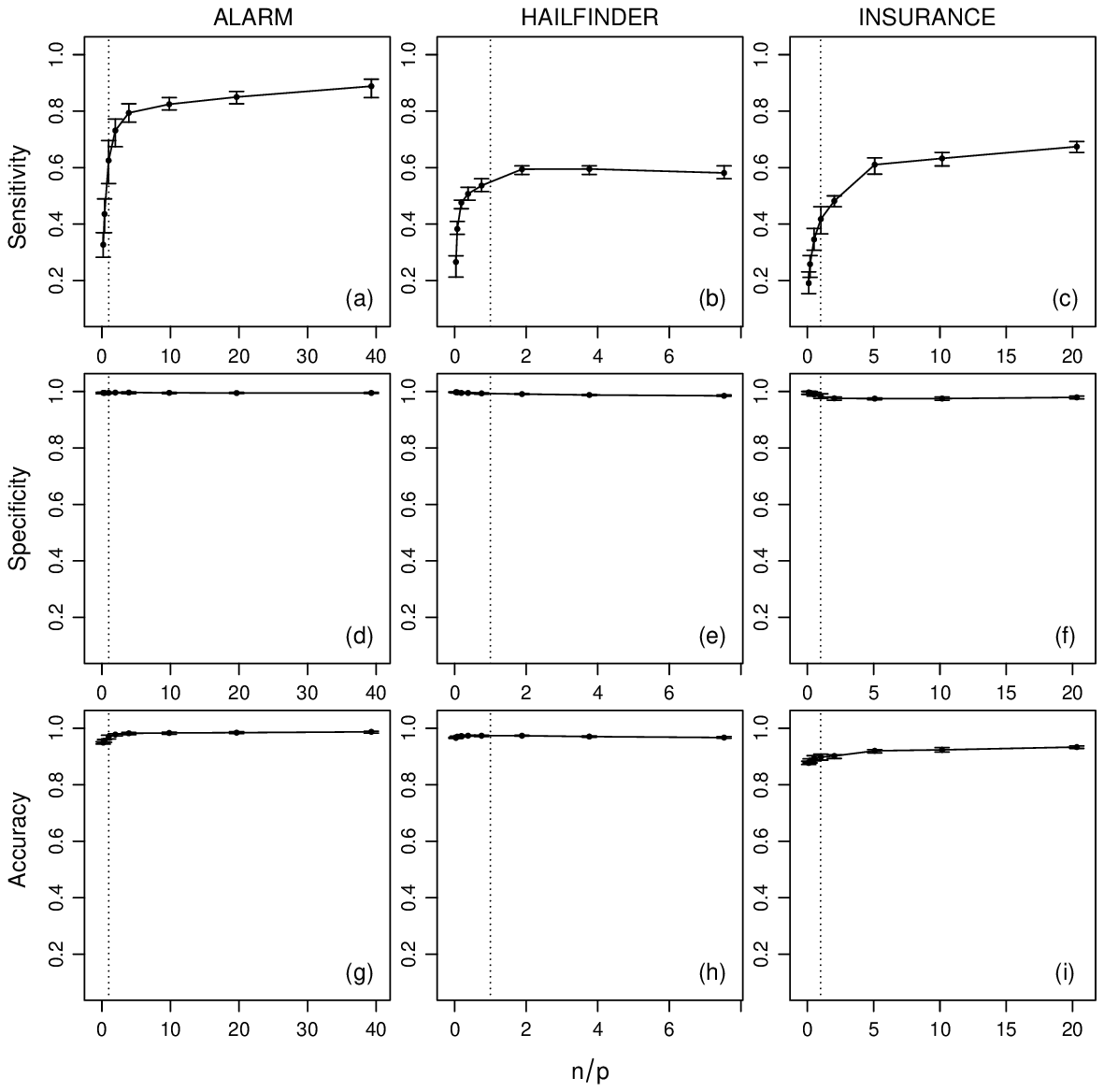}
  \caption{Average sensitivity, specificity and accuracy of MMHC for the ALARM, 
    HAILFINDER and INSURANCE networks over $n/p$. Bars represent 95\% confidence
    intervals, and the dotted vertical line is $n = p$.}
  \label{fig:roc-mmhc}
\end{center}
\end{figure}

The average values of sensitivity, specificity, accuracy and $\hat t$ for the 
networks across various sample sizes ($n$) are shown in Fig. \ref{fig:roc-iamb}
(IAMB), Fig. \ref{fig:roc-hc} (HC) and Fig. \ref{fig:roc-mmhc} (MMHC). Since
the number of parameters is non-constant across the networks, a normalised ratio
of the size of the generated sample to the number of parameters of the network
(i.e. $n/p$) is used as a reference instead of the raw sample size (i.e. $n$).
Intuitively, a sample of size of $n = 1000$ may be large enough to estimate
reliably a small network with few parameters, say $p = 100$, but it may be too
small for a larger network with $p = 10000$. On a related note, denser networks
(i.e. networks with a large number of edges compared to the number of nodes)
usually have a higher number of parameters than sparser ones (i.e. networks
with few edges).

Several interesting trends emerge from the estimated quantities. As expected,
sensitivity increases as the sample size grows. This provides an empirical
verification that the combination of HC and BDe is indeed consistent, as proved
by Chickering \cite{ges}. No analogous result exists for IAMB or MMHC, although
intuitively their sensitivity should improve as well with the sample size due
to the consistency of the conditional independence tests used by those algorithms.
Moreover, even when $n/p$ is extremely low a substantial proportion of the 
network structure can be correctly identified. When $n/p$ is at least $0.2$
(i.e. $1$ observation every $5$ parameters), HC successfully recovers from 
about $50\%$ (for ALARM and INSURANCE) to $75\%$ (for HAILFINDER) of the true
network structure. In contrast, IAMB and MMHC successfully recover from
about $45\%$ to $50\%$ of HAILFINDER, but only about $26\%$ to $40\%$ of ALARM
and $19\%$ to $30\%$ of INSURANCE. This difference in performance can be
attributed to the sparsity-inducing effect of shrinkage tests \cite{pesarin10},
which increase specificity at the cost of sensitivity. For values of $n/p$
greater than $1$ (i.e. more observations than parameters) the increase in
sensitivity slows down for all combinations of networks and algorithms,
reaching a plateau. 

Overall, sensitivity seems to have an hyperbolic behaviour, growing very rapidly
for $n/p \leqslant 1$ and then converging asymptotically to $1$ for $n/p > 1$.
Thus we expect it to increase linearly on a $\log(n/p)$ scale. The slower 
convergence rate observed for the INSURANCE network compared to the other two
networks is likely to be a consequence of its high edge density ($1.92$ edges
per node) relative to ALARM ($1.24$) and HAILFINDER ($1.17$). Slower convergence
may also be an outcome of inherent limitations of structure learning algorithms
in the case of dense networks \cite{koller,egs}.

Furthermore, both specificity and accuracy are close to $1$ for all the networks
and the sample sizes considered in the analysis, even at very low $n/p$ ratios.
Such high values are a result of the low number of true edges in ALARM, HAILFINDER
and INSURANCE compared to the respective numbers of possible edges. This is true
in particular for the ALARM and HAILFINDER networks. The lower values observed
for the INSURANCE network can be attributed again to the inherent limitations of
structure learning algorithms in modelling dense networks. The sparsity-inducing
effect of shrinkage tests is again evident for both IAMB and MMHC; both 
specificity and accuracy actually decrease slightly as $n/p$ grows and the
influence of shrinkage decreases.
 
\begin{figure}[t!]
    \includegraphics[width=\textwidth]{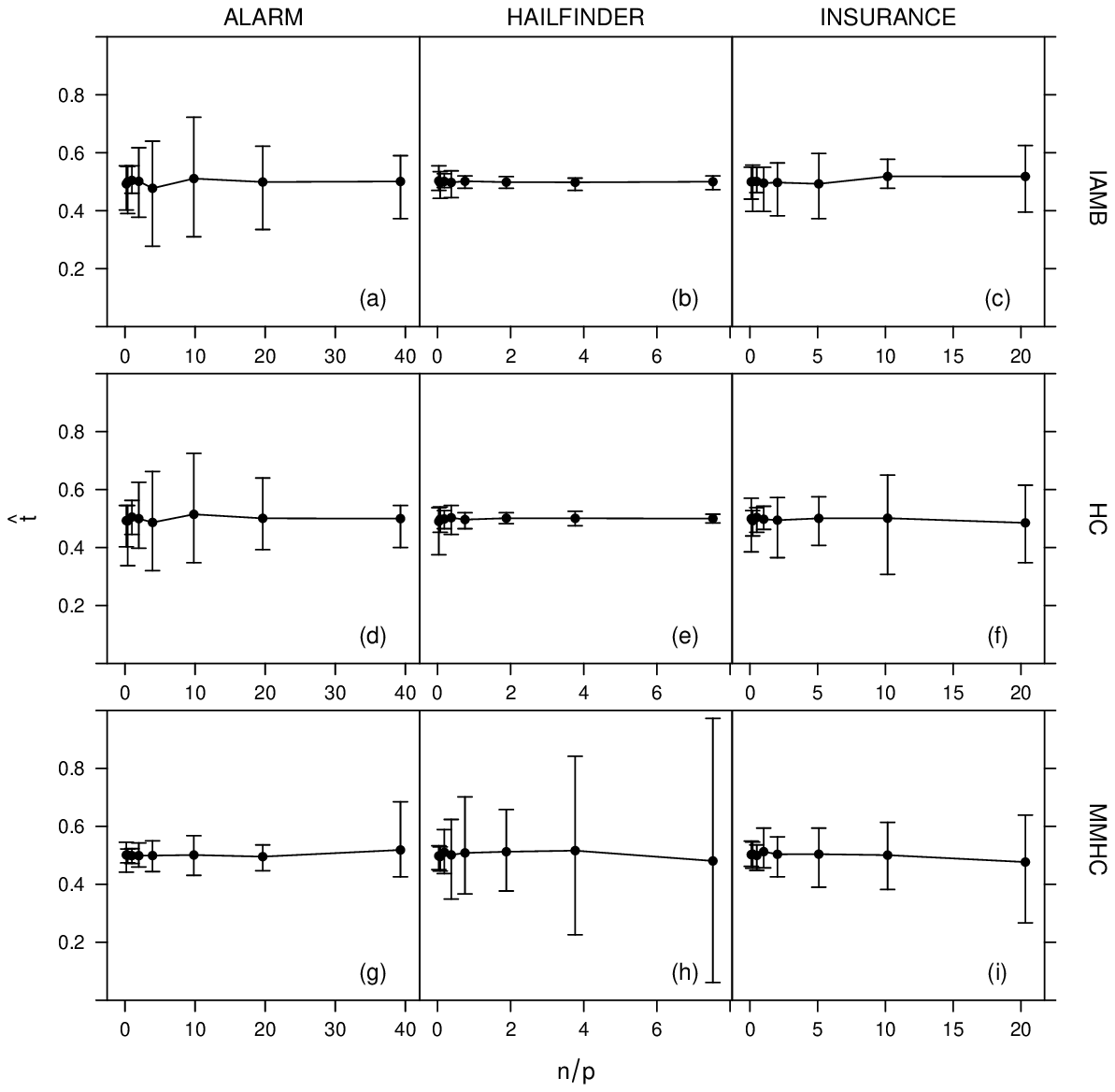}
    \caption{Average  estimated significance threshold ($\hat t$) for the ALARM,
      HAILFINDER and INSURANCE networks over $n/p$. Bars represent 95\% 
      confidence intervals.}
  \label{fig:thr}
\end{figure}

It is also important to note that, as shown in Fig. \ref{fig:thr}, the average
value of the confidence threshold $\hat t$ does not exhibit any apparent trend
as a function of $n/p$. In addition, its variability does not appear to decrease
as $n/p$ grows. This suggests that the optimal $\hat t$ depends strongly on the
specific sample used in the estimation of the confidence values $\mathbf{\hat{p}}$,
even for relatively large samples. However, specificity, sensitivity and accuracy
estimates appear on the other hand to be very stable (all confidence intervals
shown in Fig. \ref{fig:roc-iamb}, Fig. \ref{fig:roc-hc} and Fig. \ref{fig:roc-mmhc}
are very small).

\begin{figure}[t!]
\begin{center}
  \includegraphics[width=\textwidth]{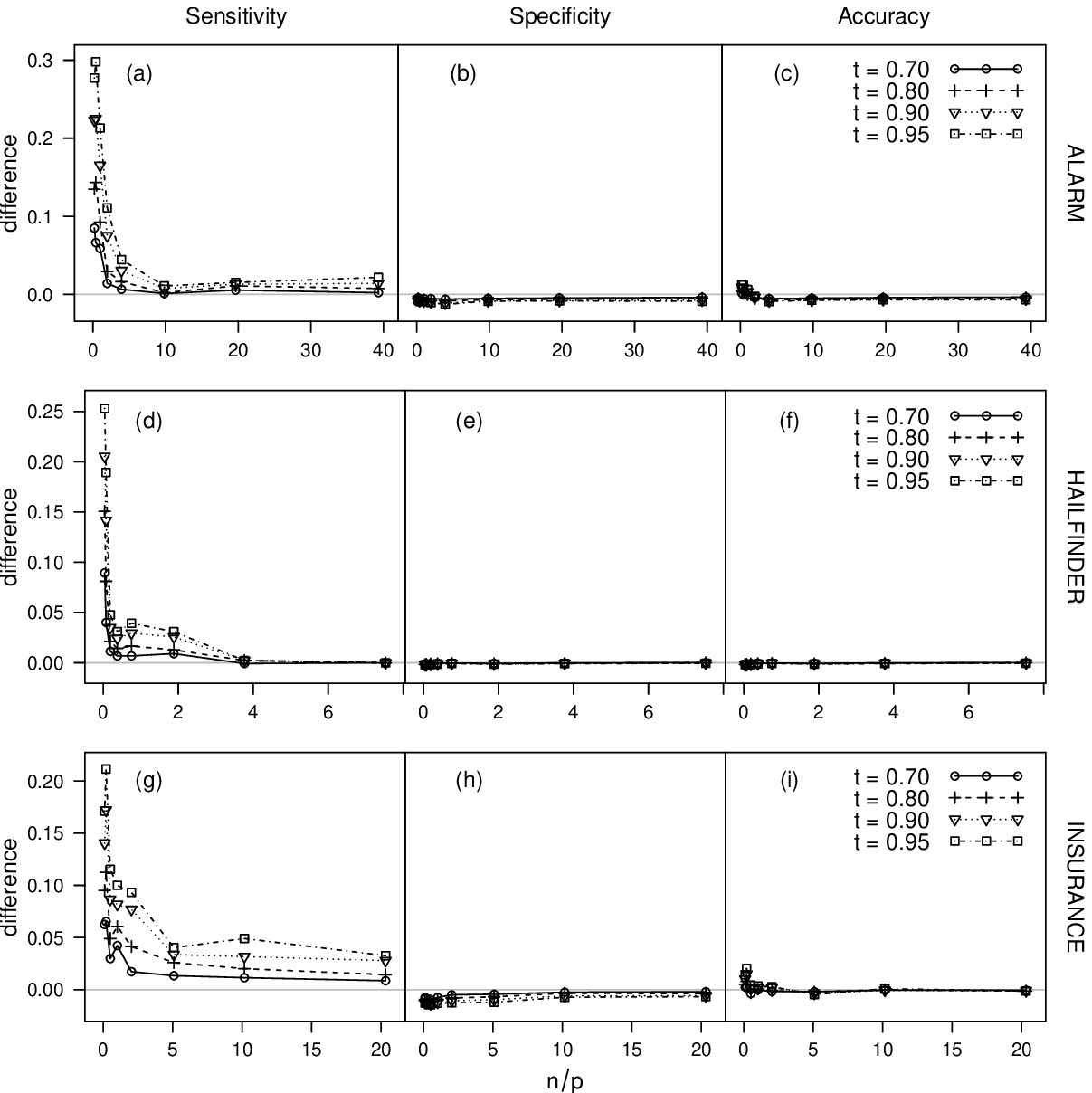}
  \caption{Difference in sensitivity, specificity and accuracy between the estimated
    threshold $\hat t$ and several ad-hoc ones ($t = 0.70, 0.80, 0.90, 0.95$) for HC
    over $n/p$.}
  \label{fig:deltas}
\end{center}
\end{figure}

From Fig. \ref{fig:thr}, it is also apparent that the threshold estimate $\hat t$
can be significantly lower than $1$ even for high values of $n/p$. This behaviour
is observed consistently across the three networks (ALARM, HAILFINDER, INSURANCE).
These results are in sharp contrast with ad-hoc thresholds commonly found in 
literature, which are usually large \citep[e.g. 0.8 in][]{friedman}. A large 
threshold can certainly be useful in excluding noisy edges, which may result from
artefacts at the measurement and dynamical levels and from finite sample-size
effects. However, while a large ad-hoc threshold can certainly minimise false
positives, it is also expected to accentuate false negatives. Such a conservative
choice can have a profound impact on the network topology, resulting in artificially
sparse networks. The threshold estimator introduced in Sec. \ref{sec:approach}
achieves a good trade-off between incorrectly identifying noisy edges as significant
and disregarding significant ones. As an example, the difference in sensitivity,
specificity and accuracy between the estimated threshold $\hat t$ and several
large, ad-hoc ones ($t = 0.70, 0.80, 0.90, 0.95$) for HC is shown in Fig.
\ref{fig:deltas} (the corresponding plots for IAMB and MMHC are similar, and are
omitted for brevity). The threshold $\hat t$ systematically outperforms the
ad-hoc thresholds in terms of sensitivity, in particular for low values of $n/p$.
The difference progressively vanishes as $n/p$ grows. All thresholds have
comparable levels of specificity and accuracy.

On a related note, false negatives across ad-hoc thresholds may also be
attributed to the fact that edges are considered as separate, independent entities
as far as the choice of the threshold is concerned -- i.e. a $0.99$ threshold is
expected to identify as significant about $1$ in $100$ edges in the network. 
However, in a biological setting the structure of the network is an abstraction
for the underlying functional mechanisms; as an example, consider the signalling
pathways in a transcriptional network. In such a context, edges are clearly not
independent, but appear in concert along signalling pathways. This interdependence
is accounted for in the proposed approach (that is based on the full set
$\mathbf{\hat p}$ of estimated conﬁdence values), but it is not commonly
considered in choosing ad-hoc thresholds.  For instance, edges appearing with
individual confidence values far below the $[0.80, 1]$ range may not necessarily
be identified as significant by an ad-hoc threshold. However, the proposed
approach recognises their interplay and correctly identifies them as significant.
This aspect, along with the strong dependence between the optimal $\hat t$ and
the actual sample the network is learned from, may discourage the use of an a
priori or ad-hoc confidence threshold in favour of more statistically-motivated
alternatives.

\section{Applications to molecular expression profiles}
\label{sec:geneprof}

In order to demonstrate the effectiveness of the proposed approach on experimental
data sets, we will examine two gene expression data sets from Nagarajan et al.
\cite{myogenic} and Sachs et al. \cite{sachs}. All the analyses will be performed
again with the bnlearn package. Following Imoto et al. \cite{imoto}, we will 
consider the edges of the BNs disregarding their direction when
determining their significance. Edges identified as significant will then be
oriented according to the direction observed with the highest frequency in the
bootstrapped networks $\mathcal{G}_b$. While simplistic, this combined approach
allows the proposed estimator to handle the edges whose direction cannot be
determined by the structure learning algorithm possibly due to score equivalent
structures \cite{chickering}.

\subsection{Differentiation potential of aged myogenic progenitors}

In a recent study \cite{myogenic} the interplay between crucial myogenic 
(Myogenin, Myf-5, Myo-D1), adipogenic (C/EBP$\alpha$, DDIT3, FoxC2, PPAR$\gamma$),
and Wnt-related genes (Lrp5, Wnt5a) orchestrating aged myogenic progenitor
differentiation was investigated by Nagarajan et al. using clonal gene expression
profiles in conjunction with BN structure learning techniques.
The objective was to investigate possible functional relationships between
these diverse differentiation programs reflected by the edges in the resulting
networks. The clonal expression profiles were generated from RNA isolated across
34 clones of myogenic progenitors obtained across 24-month-old mice and real-time
RT-PCR was used to quantify the gene expression. Such an approach implicitly
accommodates inherent uncertainty in gene expression profiles and justified the
choice of probabilistic models.

In the same study, the authors proposed a non-parametric resampling approach to
identify significant functional relationships. Starting from Friedman's definition
of confidence levels (Eq. \ref{eq:bootconf}), they computed the \emph{noise floor
distribution} $\mathbf{\hat{f}} = \{ \hat{f}_1, \hat{f}_2, \ldots, \hat{f}_k \}$
of the edges by randomly permuting the expression of each gene and performing
BN structure learning on the resulting data sets. An edge $e_i$
was deemed significant if $\hat{p}_i > \max_{\{\hat{f}_l \in \mathbf{\hat{f}}\}}
\hat{f}_l$. In addition to revealing several functional relationships documented
in literature, the study also revealed new relationships that were immune to the
choice of the structure learning techniques. These results were established across
clonal expression data normalised using three different housekeeping genes and
networks learned with three different structure learning algorithms.

\begin{figure}[t]
  \includegraphics[width=\textwidth]{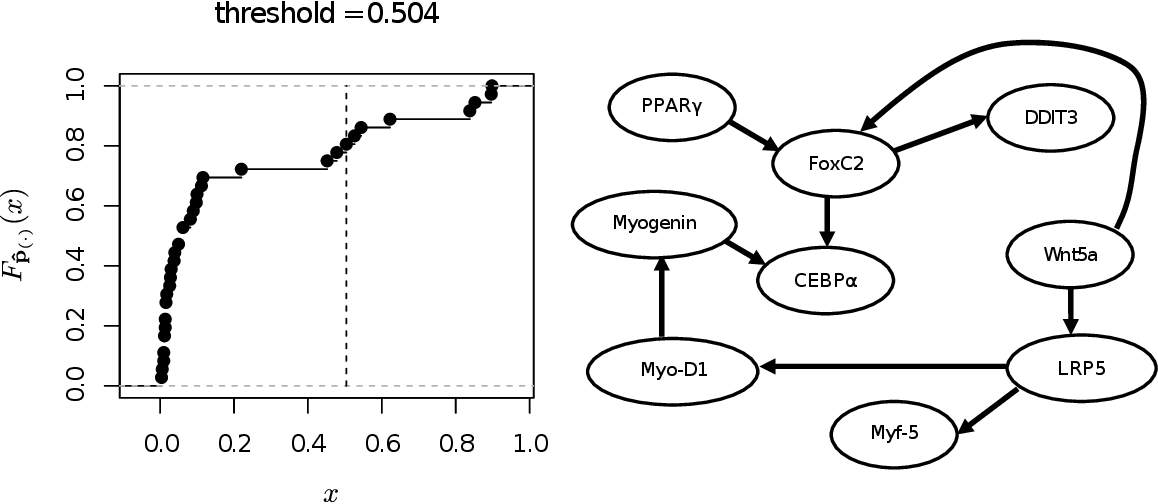}
  \caption{The empirical CDF $F_{\mathbf{\hat{p}_{(\cdot)}}}$ for the myogenic
    progenitors data from Nagarajan et al. \cite{myogenic} (on the left), and
    the network structure resulting from the selection of the significant edges
    (on the right). The vertical dashed line in the plot of
    $F_{\mathbf{\hat{p}_{(\cdot)}}}$ represents the threshold
    $F^{-1}_{\mathbf{\tilde{p}_{(\cdot)}}}(\hat{t})$.}
  \label{fig:myogenic}
\end{figure}

The approach presented in \cite{myogenic} has two important limitations. First,
the computational cost of generating the noise floor distribution may discourage
its application to large data sets. In fact, the generation of the required
permutations of the data and the subsequent structure learning (in addition to
the bootstrap resampling and the subsequent learning required for the estimation
of $\mathbf{\hat{p}}$) essentially doubles the computational complexity of
Friedman's approach. Second, a large sample size may result in an extremely low
value of $\max(\mathbf{\hat{f}})$, and therefore in a large number of false
positives.

In the present study, we re-investigate the myogenic progenitor clonal expression
data normalised using housekeeping gene GAPDH with the approach outlined in Sec.
\ref{sec:approach} and the IAMB algorithm. It is important to note that this
strategy was also used in the original study \cite{myogenic}, hence its choice.
The order statistic $\mathbf{\hat{p}_{(\cdot)}}$ was computed from $500$
bootstrap samples. The empirical CDF $F_{\mathbf{\hat{p}_{(\cdot)}}}$, the 
estimated threshold and the network with the significant edges are shown in Fig.
\ref{fig:myogenic}. 

All edges identified as significant in the earlier study \cite{myogenic} 
across the various structure learning techniques and normalisation techniques
were also identified by the proposed approach (see Fig. 3D in \cite{myogenic}).
In contrast to Fig. \ref{fig:myogenic}, the original study using IAMB and
normalisation with respect to GAPDH alone detected a considerable number of
additional edges (see Fig. 3A in \cite{myogenic}). Thus it is quite possible
that the approach proposed in this paper reduces the number of false positives
and spurious functional relationships between the genes. Furthermore, the
application of the proposed approach in conjunction with the algorithm from
Imoto et al. \cite{imoto} reveals directionality of the edges, in contrast
to the undirected network reported by Nagarajan et al. \cite{myogenic}.

\subsection{Protein signalling in flow cytometry data}
\label{sec:sachs}

In a landmark study, Sachs et al. \cite{sachs} used BNs for
identifying causal influences in cellular signalling networks from simultaneous
measurement of multiple phosphorylated proteins and phospholipids across single
cells. The authors used a battery of perturbations in addition to the
unperturbed data to arrive at the final network representation. A greedy search
score-based algorithm that maximises the posterior probability of the network
\cite{heckerman} and accommodates for variations in the joint probability 
distribution across the unperturbed and perturbed data sets was used to identify
the edges \cite{cscore}. More importantly, significant edges were selected using
an arbitrary significance threshold of $0.85$ (see Fig. 3, \cite{sachs}). A
detailed comparison between the learned network and functional relationships
documented in literature was presented in the same study.

\begin{figure}[t]
  \includegraphics[width=\textwidth]{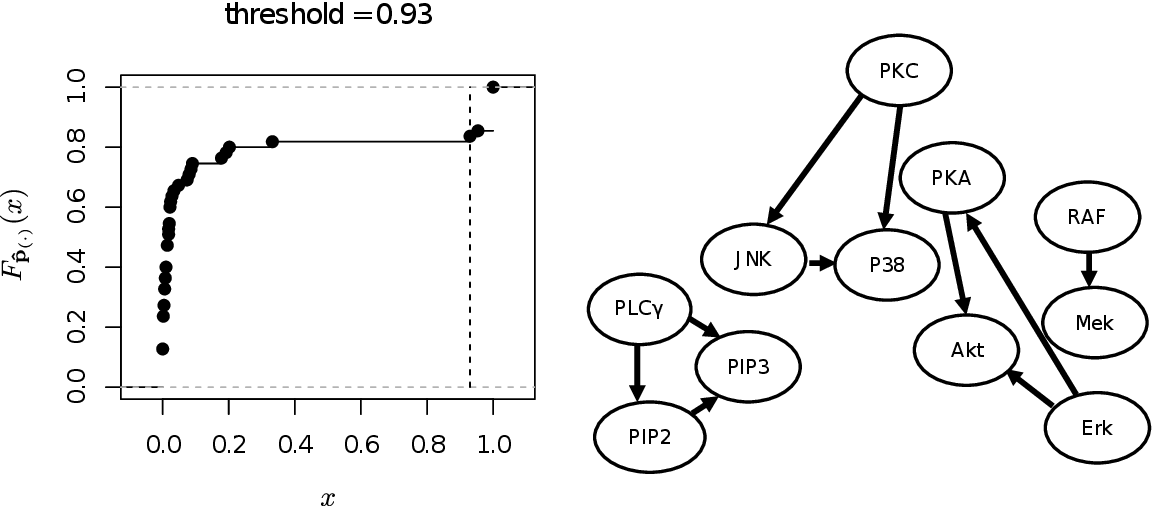}
  \caption{The empirical CDF of $\mathbf{\hat{p}_{(\cdot)}}$ for the flow 
    cytometry data from Sachs et al. \cite{sachs} (on the left), and the
    network structure resulting from the selection of the significant edges
    (on the right). The vertical dashed line in the plot of
    $F_{\mathbf{\hat{p}_{(\cdot)}}}$ represents the threshold
    $F^{-1}_{\mathbf{\tilde{p}_{(\cdot)}}}(\hat{t})$.}
  \label{fig:sachs}
\end{figure}

We investigated the performance of the proposed approach in identifying significant
functional relationships from the same experimental data. However, we limit
ourselves to the data recorded without applying any molecular intervention,
which amount to $854$ observations for $11$ variables. We compare and contrast
our results to those obtained using an arbitrary threshold of 0.85. The combination
of perturbed and non-perturbed observations studied in Sachs et al. \cite{sachs}
cannot be analysed with our approach, because each subset of the data follows a
different probability distribution and therefore there is no single ``true''
network $\mathcal{G}_0$. Analysis of the unperturbed data using the approach
presented in Sec. \ref{sec:approach} reveals the edges reported in the original
study. The resulting network is shown in Fig. \ref{fig:sachs} along with
$F_{\mathbf{\hat{p}_{(\cdot)}}}$ and the estimated threshold. From the plot of
$F_{\mathbf{\hat{p}_{(\cdot)}}}$ we can clearly see that significant and
non-significant edges present widely different levels of confidence, to the point
that any threshold between $0.4$ and $0.9$ results in the same network structure.
This, along with the value of the estimated threshold ($\hat{p}_{(i)} \geqslant
0.93$), shows that the noisiness of the data relative to the sample size is low.
In other words, the sample is big enough for the structure learning algorithm to
reliably select the significant edges. The edges identified by the proposed method
were the same as those identified by \cite{sachs} using general stimulatory cues
excluding the data with interventions (see Fig. 4A in \cite{sachs}, Supplementary
Information). In contrast to \cite{sachs}, using Imoto et al. \cite{imoto} approach
in conjunction with the proposed thresholding method we were able to identify
the directions of the edges in the network. The directions correlated with the
functional relationships documented in literature (Tab. 3, \cite{sachs},
Supplementary Information) as well as with the directions of the edges in the
network learned from both perturbed and unperturbed data (Fig. 3, \cite{sachs}).

\section{Conclusions}

Graphical models and network abstractions have enjoyed considerable attention
across the biological and medical communities. Such abstractions are especially
useful in deciphering the interactions between the entities of interest from
high-throughput observational data. Classical techniques for identifying
significant edges in the resulting graph rely on ad-hoc thresholding of the edge
confidence estimated from across multiple independent realisations of networks
learned from the given data. Large ad-hoc threshold values are particularly common,
and are chosen in an effort to minimise noisy edges in the resulting network.
While useful in minimising false positives,  such a choice can accentuate false
negatives with pronounced effect on the network topology. The present study
overcomes this caveat by proposing a more straightforward and 
statistically-motivated approach for identifying significant edges in a graphical
model. The proposed estimator minimises the $L_{\mathrm{1}}$ norm between the
CDF of the observed confidence levels and the CDF of their asymptotic, ideal
configuration. The effectiveness of the proposed approach is demonstrated on
three synthetic data sets \citep{alarm,hailfinder, insurance} and on gene
expression data sets across two different studies \cite{myogenic,sachs}. However,
the approach is defined in a more general setting and can be applied to many
classes of graphical models learned from any kind of data. 

\section*{Acknowledgements}

This work was supported by the UK Technology Strategy Board (TSB) and Biotechnology
\& Biological Sciences Research Council (BBSRC), grant TS/I002170/1 (Marco 
Scutari) and the National Library of Medicine, grant R03LM008853 (Radhakrishnan
Nagarajan). Marco Scutari would also like to thank Adriana Brogini for proofreading
the paper and providing useful suggestions.

\section*{References}

\end{document}